\documentclass[10pt,twocolumn,letterpaper]{article}

\usepackage{iccv}
\usepackage{times}
\usepackage{epsfig}
\usepackage{graphicx}
\usepackage{amsmath}
\usepackage{amssymb}
\usepackage{booktabs}
\usepackage{caption} 
\usepackage[numbers]{natbib}

\usepackage{multirow,array, bigdelim, makecell, booktabs}
\usepackage{bbding}
\usepackage{pifont}
\DeclareRobustCommand{\method}{{PAT}}
\DeclareRobustCommand{\methoddescrip}{Position-Aware Transformer}
\DeclareRobustCommand{\char}{Charades}
\DeclareRobustCommand{\thum}{MultiTHUMOS}
 
\usepackage[pagebackref=true,breaklinks=true,letterpaper=true,colorlinks,bookmarks=false]{hyperref}

\iccvfinalcopy 

\def\iccvPaperID{****} 

\ificcvfinal\fi

\begin{document}

\title{PAT: Position-Aware Transformer for Dense Multi-Label Action Detection}

\author{Faegheh Sardari$^{1}$
\and
Armin Mustafa$^{1}$
\and
Philip J. B. Jackson$^{1}$
\and
Adrian Hilton$^{1}$
\and
$^{1}$ Centre for Vision, Speech and Signal Processing (CVSSP)\\
University of Surrey, UK \\
{\tt\small \{f.sardari,armin.mustafa,p.jackson,a.hilton\}@surrey.ac.uk}
}

\maketitle
\ificcvfinal\thispagestyle{empty}\fi

\begin{abstract}
We present {\method}, a transformer-based network that learns complex temporal co-occurrence action dependencies in a video by exploiting multi-scale temporal features.  {In existing methods, the self-attention mechanism in transformers loses the temporal positional information, which is essential for robust action detection.} To address this issue, we (i) embed relative positional encoding in the self-attention mechanism and (ii) exploit multi-scale temporal relationships by designing a novel non-hierarchical network, in contrast to the recent transformer-based approaches that use a hierarchical structure. We argue that joining the self-attention mechanism with multiple sub-sampling processes in the hierarchical approaches results in increased loss of positional information. We evaluate the performance of our proposed approach on two challenging dense multi-label benchmark datasets, {and show that {\method} improves the current state-of-the-art result by $1.1\%$ and $0.6\%$ mAP on the {\char} and {\thum} datasets, respectively, thereby achieving the new state-of-the-art mAP at $26.5\%$ and $44.6\%$, respectively.} We also perform extensive ablation studies to examine the impact of the different components of our proposed network.
\end{abstract}

\section{Introduction}
\label{sec:intro}
\begin{figure}[t]
  \centering
  \includegraphics[width=1.0\linewidth]{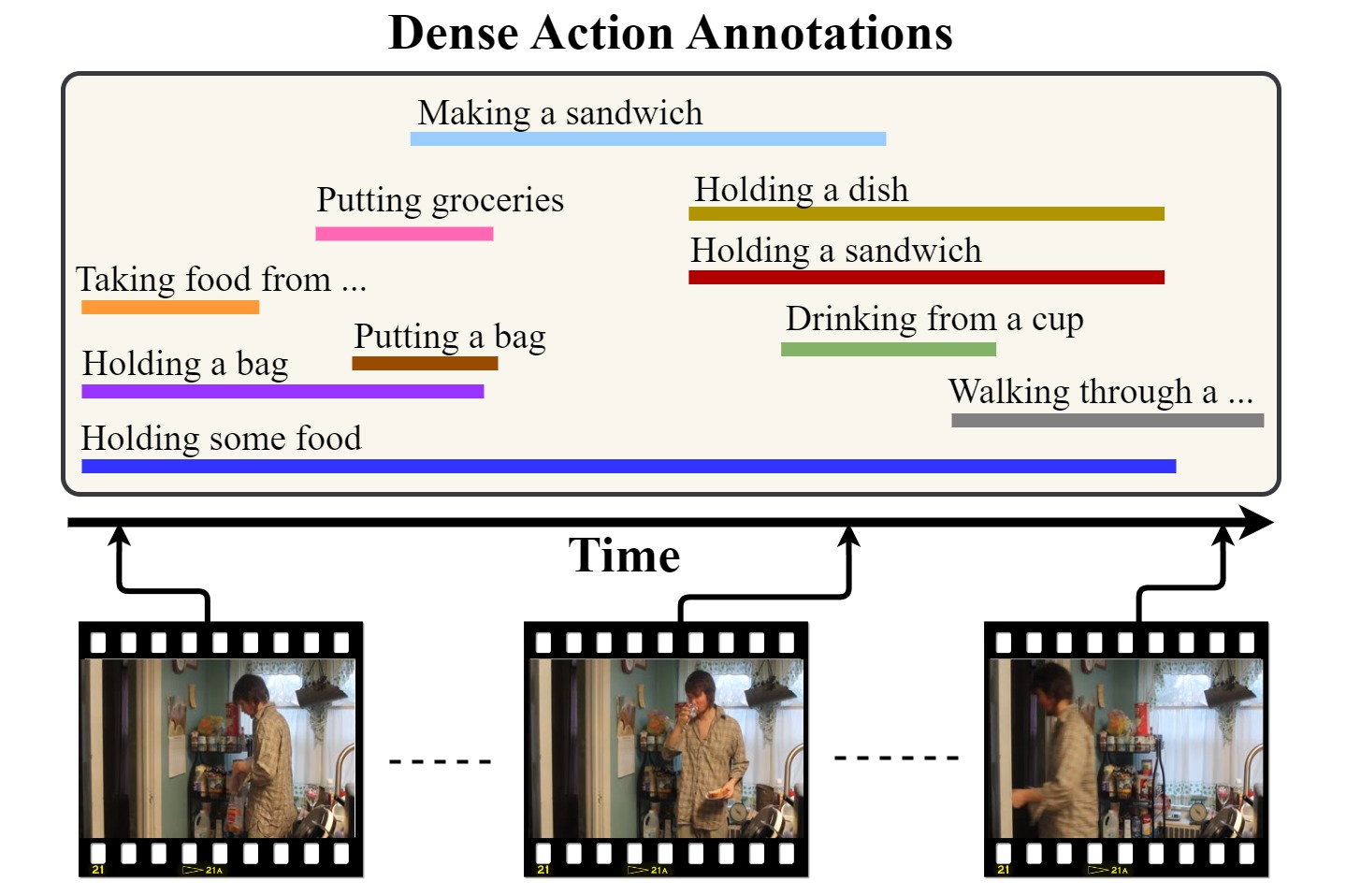}
  \caption{A sample video and its corresponding action annotations from the {\char} dataset \cite{charades} where the video includes several action types with different time spans, from short to long, and in each time step, multiple actions can occur at the same time.}
  \label{fig:dense detection}
\end{figure}

Action or event detection aims to determine the boundaries of different actions/events occurring in an untrimmed video, and plays a crucial role in various important computer vision applications, such as video summarization, highlighting, and captioning. Despite the recent advances in different areas of video understanding, dense multi-label action detection is still an unsolved problem and considered as one of the most challenging video analysis tasks since the videos are untrimmed, and include several actions with different time durations that can have overlap (See Fig. \ref{fig:dense detection}). To carry out this task, we require to learn complex short and long term {temporal relationships amongst different actions} in a video which is a challenging problem \cite{dai2022ms, kahatapitiya2021coarse}.

Most previous dense multi-label action detection approaches capture temporal dependencies through temporal convolutional networks \cite{piergiovanni2019temporal, feichtenhofer2020x3d, kahatapitiya2021coarse}. However, with the success of transformer networks over the convolutional networks for modeling complex and sequential relationships \cite{vaswani2017attention, devlin2018bert, dosovitskiy2020image,liu2021swin,wang2021pyramid}, recently, a few methods, such as \cite{tirupattur2021modeling, dai2021ctrn, dai2022ms}, leverage the self-attention mechanism and propose transformer-based approaches where they achieve state-of-the-art performance. Authors in \cite{tirupattur2021modeling, dai2021ctrn} design their network by modeling explicitly temporal cross-class relations. In \cite{tirupattur2021modeling}, there are two transformer modules such that one of them investigates action relationships for each temporal moment, and another one learns temporal dependencies for each action type. However, these approaches are not computationally efficient and their complexity grows with the number of action classes. To overcome this, \citet{dai2022ms} design a hierarchical network that learns temporal-action dependencies from multi-scale temporal features. Their network contains several transformer layers such that the output of each layer is down-sampled and given as input into its subsequent layer. As stated in \cite{shen2018disan, li2021learnable, dufter2022position}, the self-attention mechanism in the transformer is order-invariant and loses positional information, and when the self-attention is embedded in a hierarchical structure, the issue becomes worse as using multiple down-sampling processes results in {increased loss of} positional information, especially in top layers. {In this paper, we tackle these issues by introducing {\method}, a position-aware transformer network for dense action detection. {\method} consists of three main modules: fine detection, coarse detection, and classification. The fine detection module learns fine-grained action dependencies from the full temporal resolution of the video sequence for coarse detection and classification modules. The coarse detection module captures various ranges of coarse action dependencies from the fine-grained features using a non-hierarchical structure, which preserves the positional information. To further leverage the positional information, {\method} incorporates a learnable relative positional encoding \cite{shaw2018self} in the transformer layers of both fine and coarse detection models. Finally, the classification module estimates the probabilities of different action classes for every timestamp in the input video using both fine and coarse-grained action dependencies. Our key contributions can be summarized as follows:}
\begin{itemize}
    \item {For the first time, we introduce the idea of leveraging positional information in transformers for action detection} 
    \item {We design a novel {non-hierarchical} transformer-based network that preserves positional information when learning multi-scale temporal action dependencies}
    \item {We evaluate the proposed method's performance on two challenging benchmark dense action detection datasets where we outperform the current state-of-the-art result by $1.1\%$ and $0.6\%$ per-frame mean average precision (mAP) on {\char} and {\thum} respectively, thereby achieving the new state-of-the-art mAP at $26.5\%$ and $44.6\%$, respectively}
    \item {We perform extensive ablation studies to evaluate {our network design}}
\end{itemize}

\section{Related Works}
\label{sec:related works}
Although action detection \cite{chao2018rethinking, lin2019bmn, lin2021learning,lin2019bmn, ma2020sf, zhang2022actionformer, vahdani2022deep,xu2020g, chang2021augmented} has been studied significantly in computer vision, few works \cite{piergiovanni2018learning,dai2019tan, piergiovanni2019temporal, dai2021pdan, kahatapitiya2021coarse} have explored it in a dense multi-labelled setup where instances of different actions or events can overlap in different parts of a video. In this section, we review the action detection approaches by focusing on a dense-labelled setting. 

To detect the boundaries of different actions, the authors in \cite{chao2018rethinking,lin2018bsn,liu2019multi,li2021three} propose anchor-based methods where they first generate several proposals for each frame of video by using multi-scale anchor boxes, and then refine them to achieve the final action boundaries. However, these approaches are not usually applied for a dense multi-label scenario, as to model effectively the dense action distributions, they need a large amount of anchors \cite{dai2022ms}. To overcome this, some works, such as \cite{piergiovanni2018learning,dai2019tan, piergiovanni2019temporal, dai2021pdan, kahatapitiya2021coarse}, design anchor-free approaches for dense action detection. \citet{piergiovanni2018learning} propose a network that represents an untrimmed video into multi-activity events. They design multiple temporal Gaussian filters which are applied separately on the video frame features while a soft-attention mechanism is employed to combine the output of the filters to generate a global representation. Later in \cite{piergiovanni2019temporal}, they improve their work by proposing a temporal convolutional network using Gaussian filters as kernels to perform the temporal representation in a more efficient and effective way. Although they design networks to address complex multi-label action detection, the proposed models are not able to encode long-term dependencies and mostly focus on local relationships, {while our proposed network is able to capture different ranges of temporal features from short to long.} \citet{kahatapitiya2021coarse} propose a two-stream network to capture long term information such that one of the streams learns the most informative frame of a long video through a dynamic sub-sampling with a ratio of 4, and the other one learns the fine-grained contexts of the video from the full resolution. Although their results are promising, it cannot be adapted easily to use more temporal resolutions as it requires a dedicated Convolutional Neural Network (CNN), \ie X3D \cite{feichtenhofer2020x3d}, for each resolution, {whereas in our proposed method, a different resolution can be processed easily by adding an extra branch containing a few transformer blocks in the {coarse detection module}.}

\vspace{2mm}
\noindent{{\bf Transformer-based Approaches -- }}
With the success of transformer networks in modeling complex relationships and capturing short and long term dependencies \cite{vaswani2017attention, devlin2018bert, dosovitskiy2020image,liu2021swin,wang2021pyramid}, some works, such as \cite{dai2021pdan, tirupattur2021modeling, dai2022ms}, develop transformer-based approaches for dense action detection task. \citet{tirupattur2021modeling} design a model with two transformer branches, one branch applies self-attention across all action classes for each time step to learn the relationships amongst actions, and another branch uses self-attention across {time frames} to model the temporal dependencies, and the output of two branches are combined for action classification. Although this method outperforms state-of-the-art results, the method's computational complexity increases with the number of action classes. Similar to \cite{kahatapitiya2021coarse} that benefits different temporal resolutions, \citet{dai2022ms} extract multi-scale features. They design a transformer-based hierarchical structure and provide multi-resolution temporal features through several sub-sampling processes. However, as the self-attention mechanism does not preserve the temporal position information \cite{shen2018disan, li2021learnable, dufter2022position}, joining it with multiple sub-sampling processes makes the network lose more positional information while preserving this information is essential for action detection.  {In contrast, our position-aware transformer network {\method} has been designed to retain} {such temporal cues.}

\begin{figure*}[t]
  \centering
  \includegraphics[width=0.85\linewidth]{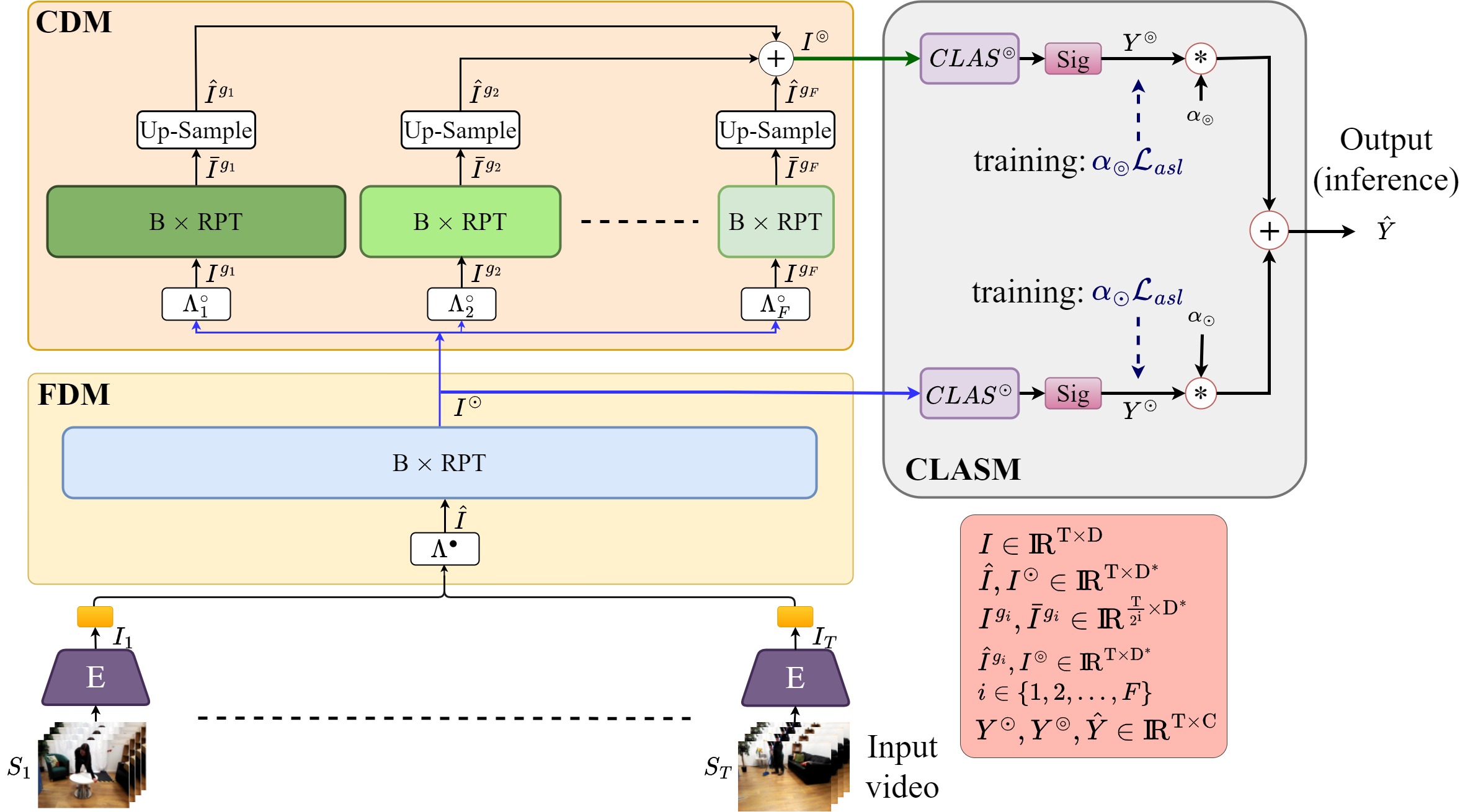}
  \caption{The overall schema of the proposed network {\method} including (i) video encoder E, (ii) fine detection module FDM, (iii) Coarse detection module CDM, and (iv) classification module CLASM.}
  \label{fig:MF-ViT}
\end{figure*}

\section{{\methoddescrip} ({\method})}
\label{sec:proposed method}
\noindent {\bf Problem Definition -- } Our aim is to detect different actions/events in a densely-labelled untrimmed video. {We define the action detection problem under this setting as \cite{kahatapitiya2021coarse, tirupattur2021modeling, dai2022ms}}. For an untrimmed video sequence with a length of $T$, each timestamp $t$ has a ground truth action label ${G}_t =\{{{g}_{t,c}}\in\{0, 1\}\}^{C}_{c=1}$, where $C$ is the maximum number of action classes in the dataset, and the network requires to estimate action class probabilities $Y_t =\{y_{t,c}\in[0, 1]\}^{C}_{c=1}$ for each timestamp.

\subsection{Proposed Network}
Our proposed method {\method} is a transformer-based network designed to exploit different granularities of complex temporal dependencies for action detection. The {\method} network includes a video encoder E that encodes an input video sequence into a sequence of input tokens, and three main components: fine detection module FDM, coarse detection module CDM, and classification module CLASM arranged as shown in Fig. \ref{fig:MF-ViT}.  FDM processes an input sequence in its original temporal resolution to obtain a fine-grained action representation for both CDM and CLASM modules. The CDM module learns different ranges of temporal action dependencies amongst the fine-grained features through extracting and combining multi-scale temporal features. CLASM estimates class probabilities from the output of both FDM and CDM modules.

\vspace{3mm}
\noindent {\bf Video Encoder (E) --} To process an input video, {\method} needs to convert it into a sequence of tokens. To perform this, similar to the previous action detection approaches \cite{tirupattur2021modeling, dai2022ms, zhang2022actionformer}, we first divide the L-frame input video $V\in \rm {I\!R}^{L\times Ch\times W \times H}$ into T non-overlapped segments $S=\{S_t\}^{T}_{t=1}$, where ${S_t}\in \rm {I\!R}^{Z\times Ch\times W \times H}$, $Z=L/T$, and $Ch$, $W$, and $H$ define number of channels, width, and height of each video frame respectively. Then, the video encoder E that is a pre-trained convolutional network is employed on each segment to {generate its corresponding token $I_t = {E}(S_t)$, where $I_t\in \rm {I\!R}^{D}$.}

\begin{figure}[t]
  \centering
  \vspace{3mm}
  \includegraphics[width=1.0\linewidth]{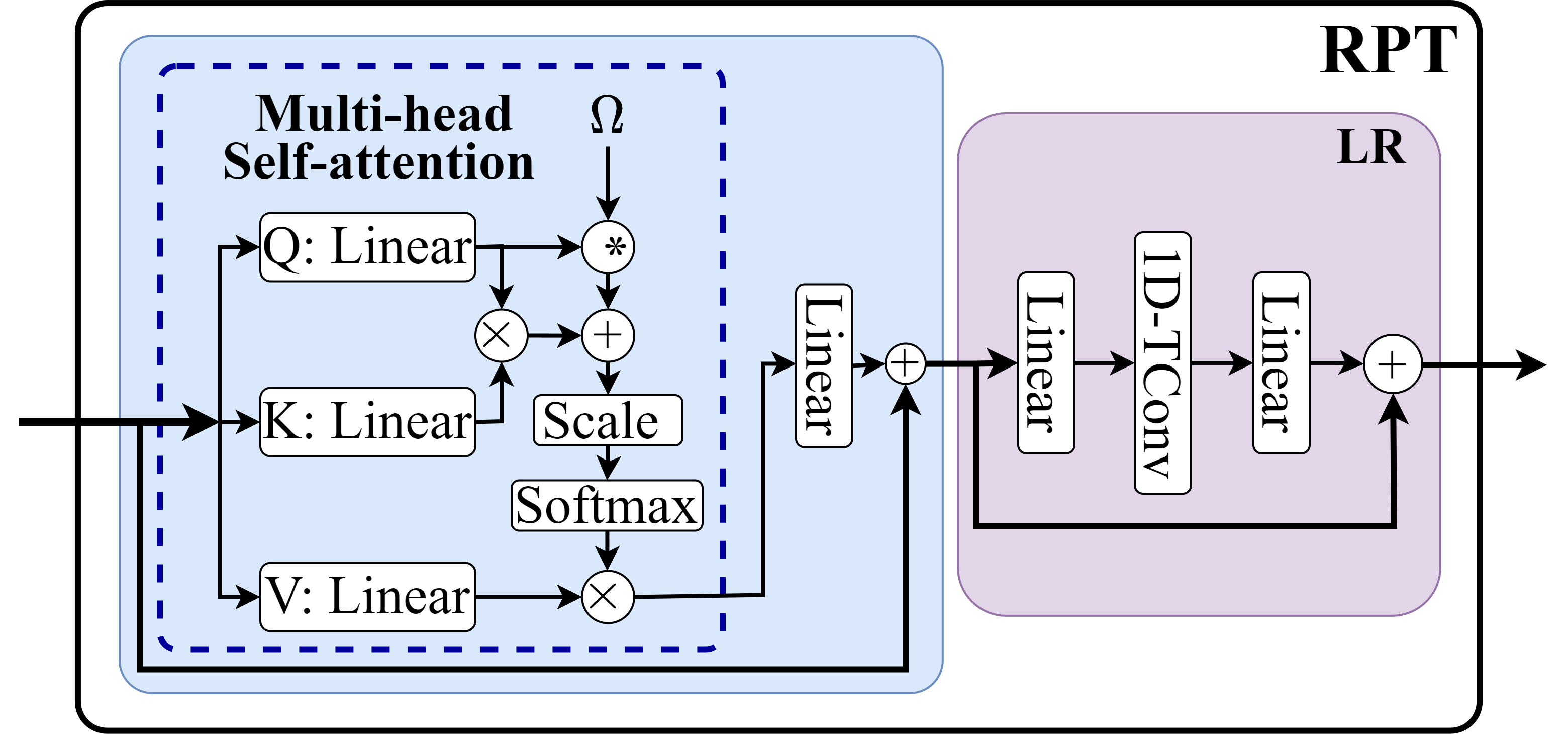}
  \caption{Architecture of {the proposed} RPT block. For brevity, the computation of the heads are not shown separately.}
  \label{fig:RPT}
\end{figure}

\vspace{3mm}
\noindent {\bf Relative Positional Transformer (RPT) Block --}
To design FDM, and CDM, we employ {our proposed} transformer block RPT (see Fig. \ref{fig:RPT}). The RPT block comprises a transformer layer with relative positional embedding followed by a local relational LR component containing two linear layers and one 1D temporal convolutional layer as in \cite{dai2022ms} to enhance the output of the transformer layer.

As already pointed out in Section \ref{sec:intro}, the transformer self-attention mechanism loses the order of temporal information while preserving this information is essential for action detection, where we need to localise events precisely in a video sequence. To solve this issue, \citet{vaswani2017attention} propose to add the absolute positional embedding to the input tokens. However, in our experiments, we observed that using the absolute positional embedding decreases the method’s performance significantly (see Section \ref{sec:ablation}). This has also been observed in \cite{dai2022ms, zhang2022actionformer}. The decrease in performance may be attributed to breaking the translation-invariant property of the method. In action detection, we expect the proposed method to be translation-invariant, \ie the network learns the same representation for the same video frames in two temporally shifted videos, regardless of how much they are shifted, while the absolute encoding can break this property as it adds different positional encodings to the same frames in the shifted video inputs. {To overcome this, we propose to use relative positional encoding \cite{shaw2018self} in the transformer layers of our RPT block.} The relative positional encoding employs a relative pairwise distance between every two tokens and is translation-invariant. In addition, as the embedding is performed in each transformer layer and is passed into the subsequent layer, the positional information can flow to the classification module where the final estimations are provided.

We briefly formulate the transformer layer in the RPT block. In the H-head self-attention layer of RPT, for each head $h \in \{1, 2, ..., H\}$, the input sequence $X\in \rm {I\!R}^{N\times D^{\diamond}}$ is first transferred into query ${Q_h}$, key ${K_h}$, and value ${V_h}$ through linear operations
\begin{equation}
{Q}_{h} = XW_h^q,~~ {K}_{h} = XW^{k}_{h},~~{V}_{h} = XW^{v}_{h},
\end{equation}
where ${Q_h}$, ${K_h}$, ${V_h}\in \rm {I\!R}^{N\times D_h}$, $W_h^q,~W_h^k,~W_h^v\in \rm {I\!R}^{D^{\diamond}\times D_h}$ refer the  weights of linear operations, and $D_h = \frac{D^{\diamond}}{H}$. Then, the self-attention with relative positional embedding is computed for each head as 
\begin{equation}
{A}_h = softmax(\frac{{Q}_h {K}^T_h+{P}^{\triangleright}_h}{\sqrt{D_h}}){V}_h,
\end{equation}
\begin{equation}
{P}^{\triangleright}_h(n,m) = \sum_{d=1}^{D_{h}} {Q}_h(n, d)\Omega_{d}(n-m),
\end{equation}
where ${{P}^{\triangleright}_h}\in \rm {I\!R}^{N\times N}$, $n,m \in\{1, 2, ..., N\}$, and $\Omega_{d}$ operates as $D_{h}$ different embeddings for time intervals based on the queries \cite{shaw2018self}. To compute ${{P}^{\triangleright}_h}$, we use the memory-efficient method proposed by \citet{huang2018music}.

Finally, the self-attention of all heads are concatenated and fed into a linear layer to output sequence ${O}$
\begin{equation}
{A} = concat(A_1, A_2,...,A_m),
\end{equation}
\begin{equation}
{O} = {A}W^{o}+X,
\end{equation}
where ${A}\in \rm {I\!R}^{N\times D^\diamond}$ and $W^{o}\in \rm {I\!R}^{D^\diamond\times D^\diamond}$.

\vspace{3mm}
\noindent {\bf Fine Detection Module (FDM) --} The FDM module aims to obtain a fine-grained temporal action dependency representation of the video from the input video sequence for the CDM and CLASM modules. FDM includes a 1D temporal convolutional layer followed by $B$ RPT blocks. The convolution layer ${\Lambda^{\bullet}}$ has a kernel size of three and a stride of one to map all the input tokens $I\in \rm {I\!R}^{T\times D}$ into a lower dimension $D^{*}$, and then the RPT blocks are applied to learn the fine-grained dependencies $I^{\odot}$.  
\begin{equation}
{I}^{\odot} = RPT^{FDM}_{1:B}({\Lambda^{\bullet}}(I)),
\end{equation}
where $I^{\odot}\in \rm {I\!R}^{T\times D^*}$ and $D^* < D$.

\begin{figure}[t]
  \centering
  \includegraphics[width=1.0\linewidth]{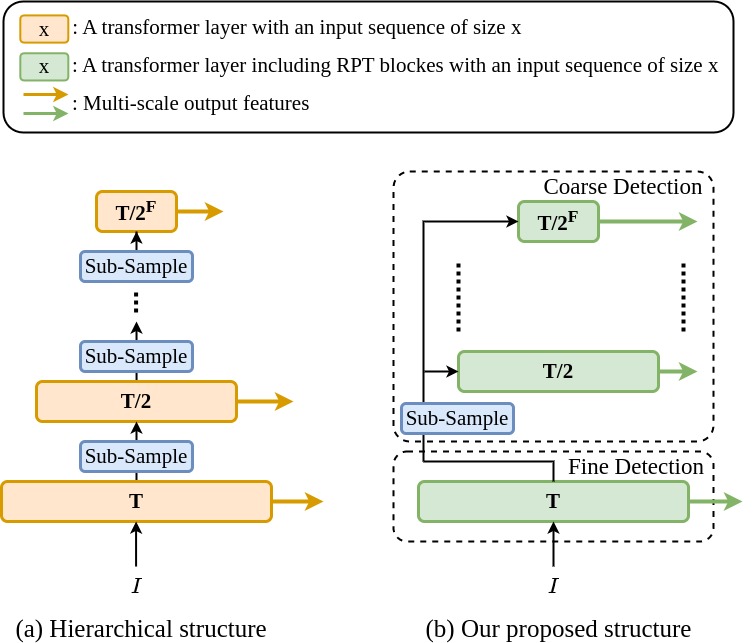}

  \caption{The proposed hierarchical structure in \cite{dai2022ms, zhang2022actionformer} vs. our proposed non-hierarchical design in fine and coarse detection modules to extract multi-scale features for action detection.}
  \label{fig:hierachical vs ours}
\end{figure}

\vspace{3mm}
\noindent {\bf Coarse Detection Module (CDM) --} In the CMD module, we aim to learn a coarse temporal action dependency representation of the video. To achieve this, one solution is to extract and combine multi-scale temporal features through a hierarchical structure, such as the proposed method in \cite{dai2022ms, zhang2022actionformer} (see Fig. \ref{fig:hierachical vs ours}. a). However, as we already explained in Section \ref{sec:intro}, using multiple sub-sampling processes in the hierarchical structure results in losing more positional information in the top layers of the network. Our CMD module has been designed to overcome this issue by extracting different scales of features from the same full-scale fine-grained information and through only one sub-sampling process, {(see Fig. \ref{fig:hierachical vs ours}. b)}. {In Section \ref{sec:ablation}, we show that our novel non-hierarchical design to extract multi-scale features outperforms significantly a hierarchical structure.} 

The CMD module has $F$ granularity branches such that each branch learns a different scale of temporal features. In the $i^{th}$ branch, first a 1D temporal convolutional layer $\Lambda^{\circ}_{i}$ with a kernel size of three and a stride of $2^i$ is applied on the fine-grained inputs received from the preceding module FCM as
\begin{equation}
{I}^{g_i} = \Lambda^{\circ}_{i}({I}^{\odot}),
\end{equation}
where ${I}^{g_i}\in \rm {I\!R}^{T^i\times D^*}$, $i\in\{1, 2, ..., F\}$, and $T^i = \frac{T}{2^i}$.
Then, the down-sampled features are given into $B$ RPT transformer blocks to exploit the temporal dependencies amongst them 
\begin{equation}
\bar{I}^{g_i} = RPT^{CDM_{i}}_{1:B}({I}^{g_i}),
\end{equation}
where ${\bar{I}}^{g_i}\in \rm {I\!R}^{T^i\times D^*}$.
Note, to extract all the scales of features, the striding process (sub-sampling) is used only once and as the relative positional information has been already embedded in the fine-grained information, after the striding process, the sub-sample features keep {the temporal positional cues}. 

In the CLASM module, action class probabilities are estimated for each input token generated by the video encoder E. Therefore, the CMD module requires obtaining a coarse dependency representation at the original temporal length. To do this, we up-sample and combine different scales of coarse features to provide a final coarse representation $I^{\circledcirc}$ as
\begin{equation}
\hat{{I}}^{g_i} = UpSample(\bar{I}^{g_i}),
\end{equation}
\begin{equation}
I^{\circledcirc} = \sum_{i=1}^{F} \hat{{I}}^{g_i},
\end{equation}
where $\hat{{I}}^{g_i}, I^{\circledcirc}\in \rm {I\!R}^{T\times D^*}$ and linear interpolation is employed for up-sampling.

\vspace{3mm}
\noindent {\bf Classification Module (CLASM) --} 
This module obtains the action class probabilities for action detection from the fine and coarse contexts. To  this, two convolution blocks $CLAS^{\odot}$  and $CLAS^{\circledcirc}$ that include two 1D convolution filters with kernel size one and stride one are applied on the fine and coarse features separately to predict $C$ action class probabilities for each temporal moment
\begin{equation}
\label{eq:class probability}
Y^{\phi} = Sig(CLAS^{\phi}(I^{\phi})),
\end{equation}
where $Y^{\phi}\in \rm {I\!R}^{T\times C}$, $\phi\in\{\odot, \circledcirc\}$, and $Sig$ refers to sigmoid activation function. Then, at inference, the final estimation is computed by combining them as
\begin{equation}
\hat{Y} = {\sum_{\phi}\alpha_{\phi}Y^{\phi}},
\end{equation}
where $\hat{Y}\in \rm {I\!R}^{T\times C}$ and $\alpha_{\odot}+~\alpha_{\circledcirc} = 1$.

\subsection{Network Optimization} 
To optimize action detection models, binary cross entropy (BCE) is usually used {as in \cite{tirupattur2021modeling, dai2022ms, zhang2022actionformer, vahdani2022deep}}. However, in the multi-label setting, the number of positive labels may become more than the number of negative ones. This unbalanced number of positive and negative labels can result in poor performance in the action detection task if we employ BCE for training, since it does not have any control on the contribution of positive and negative samples. To overcome this, {we propose to adapt Asymmetric loss $\mathcal{L}_{asl}$ \cite{ridnik2021asymmetric} for multi-label action detection}. Therfore, the total loss $\mathcal{L}_{total}$ is computed as
\begin{equation}
\mathcal{L}_{total} = \frac{1}{T}{\sum_{\phi}}\sum_{t=1}^{T}\sum_{c=1}^{C}\alpha_{\phi}\mathcal{L}_{asl}({g}_{t,c}, {y}^{\phi}_{t,c}),
\end{equation}
\begin{equation}
\mathcal{L}_{asl}({g}_{t,c}, {y}^{\phi}_{t,c}) = -{g}_{t,c}\mathcal{L}_{+}-(1-{g}_{t,c})\mathcal{L}_{-},
\end{equation}
\begin{equation}
\label{eq:focus+}
\mathcal{L}_{+} = (1-{y}^{\phi}_{t,c})^{\gamma_+}log({y}^{\phi}_{t,c}),
\end{equation}
\begin{equation}
\label{eq:focus-}
\mathcal{L}_{-} = {{(\acute{y}^{\phi}_{t,c}}})^{\gamma_-}log(1-\acute{y}_{t,c}),
\end{equation}
\begin{equation}
\label{eq:focus extra}
\acute{y}^{\phi}_{t,c} = max({y}^{\phi}_{t,c} - \delta, 0),
\end{equation}
where ${g}_{t,c}$ indicates the ground truth label of action class $c$ in temporal step $t$, and ${y}^{\phi}_{t,c}$ is its corresponding class probability estimated by Eq. \ref{eq:class probability}. $\gamma_+$ and $\gamma_-$ are focusing parameters for positive and negative labels respectively and if we choose $\gamma_+<\gamma_-$, we are able to increase the contribution of positive samples. Furthermore, Eq. \ref{eq:focus extra} applies another asymmetric mechanism by discarding the very easy negative samples through setting the threshold parameter $\delta$. In Section \ref{sec:ablation}, we show that optimizing the proposed network through Asymmetric loss instead of BCE improves the method's performance.



\section{Experimental Results}
\label{sec:experiments}
\noindent {\bf Datasets --} There are several benchmark datasets for action detection, but only a few of them provide dense multi-label annotations. For instance, videos in ActivityNet \cite{activitynet} have only one action type per timestamp. We present the results of {\method} on two challenging dense multi-label benchmark datasets, {\char} \cite{charades} and {\thum} \cite{multithomus}.

{\char} \cite{charades} is a large dataset including $9,848$ videos of daily activities of 267 persons. It contains $66,500$ temporal interval annotations for 157 action classes while there is a high overlap amongst the action instances of different action categories. To evaluate our method on {\char}, we follow previous methods \cite{kahatapitiya2021coarse,tirupattur2021modeling,dai2022ms} and use the same training and testing set as in \cite{charades}.

{\thum} contains the same set of 413 videos as in THUMOS’14 dataset \cite{jiang2014thumos}. However, \noindent {\thum} is more challenging than THUMOS’14 since (i) the annotations have been extended from 20 action classes to 65, and (ii) in contrast to sparse-lable frame-level annotations in THUMOS’14, {\thum} has dense multi-label action annotations. To obtain the results on this dataset, we use the same standard training and testing splits applied by previous methods \cite{tirupattur2021modeling, dai2022ms}. 
\noindent {Following state-of-the-art methods \cite{piergiovanni2018learning, piergiovanni2019temporal,kahatapitiya2021coarse,tirupattur2021modeling, dai2022ms}, we evaluate our method on these datasets by standard per-frame mAP metric.}

\vspace{3mm}
\noindent {\bf Implementation Details --} Similar to the proposed method in \cite{dai2022ms}, during both training and inference, {\method} uses a fixed number of $T=256$ input tokens. For training, we randomly sample a clip containing $T$ consecutive tokens from a video sequence. At inference, we follow previous work \cite{tirupattur2021modeling,kahatapitiya2021coarse} and make the predictions for a full video sequence. Each input token is provided by applying the video encoder E on an 8-frame segment to extract a feature vector with dimension $D = 1024$. The video encoder E is implemented by using a pre-trained I3D \cite{carreira2017quo}\footnote{{Video encoder E is pre-trained on Kinetic-400 \cite{kay2017kinetics} and training set of {\char} for {\thum} and {\char} respectively}.} while its fully connected layers are replaced with a global average pooling layer {and its parameters are frozen}. In the convolutional layer of FDM, the input features are mapped into $D^{*}=512$ dimensional feature vectors. Note, the feature dimension $D^{*}=512$ is fixed for the rest of the network. FDM and each granularity branch of CDM have $B=3$ RPT blocks with $H=8$ multi-head attention heads, and the number of granularity branches in CMD is set to $F=3$ {as we found that with these parameters, {\method} obtains the best performance}. The contributing factors for fine-grained ($\alpha_{\odot}$) and coarse-grained ($\alpha_{\circledcirc}$) features in the CLASM module are set {empirically} to $\{\alpha_{\odot}=0.1, \alpha_{\circledcirc}=0.9\}$ and $\{\alpha_{\odot}=0.7, \alpha_{\circledcirc}=0.3\}$ for {\char} and {\thum} respectively. In Asymmetric loss,
we use factors of $\gamma_+=1$ and $\gamma_-=3$ for the impact of positive and negative samples respectively, and threshold parameter $\delta = 0.1$, which are determined through trial and error. 

Our experiments were performed under Pytorch on an NVIDIA GeForce RTX 3090 GPU, and we trained our model using the Adam optimiser \cite{kingma2014adam} with an initial learning rate of 0.0001 and batch size 3 for 25 and 300 epochs for {\char} and {\thum} datasets respectively. The learning rate was decreased by a factor of 10 every 7 and 130 epochs for {\char} and {\thum} respectively. {Note, using different training settings for {\char} and {\thum} is due to their different size.}

\vspace{3mm}
\subsection{Ablation Studies}
\label{sec:ablation}
In this section, we examine our design decisions for the proposed network and learning paradigm.

\vspace{3mm}
\noindent {\bf Effect of FDM and CDM Modules --} 
\noindent Here, we aim to evaluate the impact of the fine and coarse detection modules (FDM and CDM) in the final results of {\method}. Table \ref{tab:mf-vit components} shows per-frame mAP on the {\char} and {\thum} datasets as we remove each or both of FDM and CDM modules. To obtain the results of the network when both modules are dropped, we use directly the sequence of input tokens generated by the video encoder (I3D) for action detection. Table \ref{tab:mf-vit components} shows that using only input tokens generated by I3D network is not enough for effective action detection and employing fine and coarse-grained temporal features obtained by FDM and CDM improves the performance by $9.7\%$ and $7.9\%$ per-frame mAP on {\char} and {\thum} respectively. It also shows that both FDM and CDM modules have an important contribution to the final results as by removing FDM and CDM, our results deteriorate by $2.4\%$ and $3.4\%$ per-frame mAP on average on both datasets respectively. Table \ref{tab:mf-vit components} also shows that for different datasets that have different action types, the contribution of fine and coarse features might be different which is the reason we use the contribution factors $\{\alpha_{\odot}, \alpha_{\circledcirc}\}$ to combine the prediction results of FDM and CDM in the CLASM module at the inference. 

\begin{table}[h]
\scalebox{1.0}
{
  \setlength{\tabcolsep}{6pt} 
    \renewcommand{\arraystretch}{1.0} 
  \begin{tabular}{@{}l c c @{}}\specialrule{.2em}{.1em}{.1em}
    \multirow{2}{*}{{Module}}& \multicolumn{2}{c}{{mAP(\%)}}\\\cmidrule{2-3}
    & {\char} & {\thum}\\ \specialrule{.2em}{.1em}{.1em}
   {CLASM} & 16.8& 36.7\\
   {FDM, CLASM} & 23.8 & 40.5\\ 
   {CDM, CLASM} & 26.2 & 40.1\\ \midrule
   {FDM, CDM, CLASM} & \bf 26.5& \bf 44.6\\ \specialrule{.2em}{.1em}{.1em}
  \end{tabular}}
  \caption{Ablation studies on FDM and CDM modules of {\method} on the {\char} and {\thum} dataset using RGB videos in terms of per-frame mAP metric.}
  \label{tab:mf-vit components}
\end{table}

\noindent {\bf Effect of Structure Design to Extract Multi-Scale Features --} In this section, we examine the design of {\method} with two other variants to capture fine-grained and coarse-grained features. In the first variant {\method-$v_1$}, the CDM module uses the hierarchical structure to extract the multi-scale features while the rest of its architecture is the same as {\method}. In the second variant {\method-$v_2$}, the CDM module has a non-hierarchical structure, the same as {\method}, but the FDM module and all granularity branches in CDM learn their features from input tokens. 

\begin{table}[t]
  \centering
  \setlength{\tabcolsep}{6pt} 
    \renewcommand{\arraystretch}{1.0} 
  \begin{tabular}{@{}l c c @{}}\specialrule{.2em}{.1em}{.1em}
    \multirow{2}{*}{{Design}}& \multicolumn{2}{c}{{mAP(\%)}}\\\cmidrule{2-3}
    & {\char} & {\thum}\\ \specialrule{.2em}{.1em}{.1em}
    {\method-$v_1$ (Hierarchical)} & 25.1 & 44.0\\ 
   {\method-$v_2$} & {26.1}& 44.2\\ \midrule
   {\method} & \bf 26.5& \bf 44.6\\ \specialrule{.2em}{.1em}{.1em}
  \end{tabular}
  \caption{Ablation studies on structure design of the proposed method on the {\char} and {\thum} datasets using RGB videos in terms of per-frame mAP metric.\vspace{2mm}} 
  \label{tab:different structures}
\end{table}

Table \ref{tab:different structures} shows that when CMD applies a hierarchical structure to learn the coarse-grained features, {\ie} {\method-$v_1$}, the method's performance drops $1.4\%$ and {$0.6\%$} on {\char} and {\thum} respectively. {This proves the contribution of our novel non-hierarchical transformer-based design which preserves positional information when exploiting the multi-scale features.} Furthermore, in case we apply a non-hierarchical CMD, {\ie} {\method} and {\method-$v_2$}, if the CMD module extracts the multi-scale features from the fine-grained context instead of the input tokens as in {\method}, we achieve the best performance at $26.5\%$ and $44.6\%$ per-frame mAP on {\char} and {\thum} respectively.

\noindent {\bf {Impact of Relative Positional Encoding --}} Table \ref{tab:positional encoding} shows the performance of {\method} when different positional encodings are applied. It can be observed that employing the relative positional encoding \cite{shaw2018self,huang2018music} embedded in the RPT block improves the method's performance by $0.3\%$ per-frame mAP on both datasets, while adding absolute positional encoding \cite{vaswani2017attention} into the input tokens deteriorates the method's performance significantly.

\begin{table}[h]
\scalebox{1.0}
{ 
  \setlength{\tabcolsep}{6pt} 
    \renewcommand{\arraystretch}{0.9} 
  \begin{tabular}{@{}l c c @{}}\specialrule{.2em}{.1em}{.1em}
    \multirow{2}{*}{{Positional Encoding}}& \multicolumn{2}{c}{{mAP(\%)}}\\\cmidrule{2-3}
    & {\char} & {\thum}\\ \specialrule{.2em}{.1em}{.1em}
   {No encoding} & 26.2& 44.3\\
   {Absolute} & 25.3 & 43.5\\\midrule
   {Relative} & \bf 26.5& \bf 44.6\\ \specialrule{.2em}{.1em}{.1em}
  \end{tabular}}
  \caption{Ablation studies on positional encoding used in {\method} on the {\char} and {\thum} dataset using RGB videos in terms of per-frame mAP metric.}
  \label{tab:positional encoding}
\end{table}

\vspace{3mm}
\noindent {\bf Impact of Loss Function --} Here, we examine the effect of BCE and Asymmetric \cite{ridnik2021asymmetric} losses for training. As shown in Table \ref{tab:different losses}, applying the Asymmetric loss \cite{ridnik2021asymmetric} to optimize {\method} improves the performance by $0.5\%$ and $0.2\%$ per-frame mAP on {\char} and {\thum} respectively. 

\begin{table}[h]
  \centering
  \setlength{\tabcolsep}{6pt} 
    \renewcommand{\arraystretch}{1.0} 
  \begin{tabular}{@{}l c c @{}}\specialrule{.2em}{.1em}{.1em}
    \multirow{2}{*}{{Loss}}& \multicolumn{2}{c}{{mAP(\%)}}\\\cmidrule{2-3}
    & {\char} & {\thum}\\ \specialrule{.2em}{.1em}{.1em}
    {BCE} & {26.0}& 44.4\\ \midrule
   {Asymmetric \cite{ridnik2021asymmetric}} & \bf 26.5& \bf 44.6\\ \specialrule{.2em}{.1em}{.1em}
  \end{tabular}
  \caption{Ablation studies on the loss function applied for training {\method} on the {\char} and {\thum} datasets using RGB videos in terms of per-frame mAP metric.} 
  \label{tab:different losses}
\end{table}

\vspace{3mm}
\noindent {\bf Discussion and Analysis --} {The ablation studies show that leveraging positional information in the transformer layers has an important contribution in the final results of the network where extracting the multi-scale temporal features through our proposed non-hierarchical design in CMD outperforms a hierarchical structure by $1.0\%$ mAP on average on both datasets ({\method} vs {\method}-v1), and embedding the relative position encoding in the RPT block improves the performance by $0.3\%$ mAP on both datasets. Our further ablations also reveal the effect of the Asymmetric loss in optimizing of {\method}  where it increases the performance by $0.3\%$ mAP on average on both datasets.}


\subsection{State-of-the-Art Comparison} In this section, we compare the performance of the proposed method with the state-of-the-art action detection approaches including both transformer-based methods and the methods that do not use self-attention. Both quantitative and qualitative results are obtained for this section. 

Table \ref{tab:sota v1} provides comparative results on the benchmark datasets {\char} and {\thum} based on the standard per-frame mAP metric. Table \ref{tab:sota v1} shows that our proposed method outperforms the current state-of-the-art result by $1.1\%$ and $0.6\%$ on {\char} and {\thum} respectively and achieves a new state-of-the-art per-frame mAP results at $26.5\%$ and $44.6\%$ on {\char} and {\thum} respectively.

\begin{table*}[h]
\begin{center}
  \centering
  \setlength{\tabcolsep}{6pt} 
    \renewcommand{\arraystretch}{1.0} 
  \begin{tabular}{@{}l l c c c c@{}}\specialrule{.2em}{.1em}{.1em}
    \multicolumn{2}{l}{\multirow{2}{*}{Method}}& \multirow{2}{*}{GFLOPs} & \multicolumn{1}{c}{\multirow{2}{*}{Backbone}} & \multicolumn{2}{c}{mAP(\%)}\\ \cmidrule{5-6}
          & & &  &\multicolumn{1}{c}{Charades} & \multicolumn{1}{c}{MultiTHUMOS}\\ \specialrule{.2em}{.1em}{.1em}
          R-C3D \cite{xu2017r}&ICCV 2017&-& C3D & 12.7 & - \\ 
          SuperEvent \cite{piergiovanni2018learning}&CVPR 2018&0.8&I3D & 18.6&36.4\\ 
          TGM \cite{piergiovanni2019temporal}&ICML 2019&1.2&I3D &20.6&37.2\\ 
          PDAN \cite{dai2021pdan}{\checkmark}$*$&WACV 2021&3.2&I3D &23.7&40.2\\ 
          CoarseFine \cite{kahatapitiya2021coarse}&CVPR 2021&-&X3D & 25.1&-\\ 
          MLAD \cite{tirupattur2021modeling}{\checkmark}&CVPR 2021&44.8 &I3D&18.4&42.2\\ 
          CTRN \cite{dai2021ctrn}{\checkmark}&BMVC 2021 &-&I3D&25.3& {{44.0}}\\
          PointTAD \cite{tanpointtad}&NeurIPS 2022&-&I3D& 21.0 & 39.8 \\
          MS-TCT \cite{dai2022ms}{\checkmark}&CVPR 2022&6.6&I3D &{{25.4}}&{43.1}\\ \midrule
          {\method}{\checkmark}& &8.5& I3D&{\bf 26.5}&{\bf 44.6} \\ \specialrule{.2em}{.1em}{.1em}
  \end{tabular}
  \caption{Action detection results on {\char} and {\thum} datasets using RGB videos in terms of per-frame mAP. The {{\checkmark}} symbol highlights the transformer-based approaches, and $*$ indicates the results are taken from \cite{dai2022ms}.\vspace{-2mm}} 
  \label{tab:sota v1}
  \end{center}
\end{table*}

\begin{table*}[h]
  \centering
  \setlength{\tabcolsep}{3.9pt} 
    \renewcommand{\arraystretch}{0.9} 
      \begin{tabular}{@{}l c c c c c c c c c c c c c c@{}} \specialrule{.2em}{.1em}{.1em}
          \multicolumn{1}{l}{\multirow{2}{*}{Method}} & \multicolumn{4}{c}{\bf {$\tau = 0$}}&\phantom & \multicolumn{4}{c}{\bf {$\tau = 20$}}&\phantom&\multicolumn{4}{c}{\bf {$\tau = 40$}}\\ \cmidrule{2-5} \cmidrule{7-10} \cmidrule{12-15}
          &{$P_{AC}$}&{$R_{AC}$}&{${F1}_{AC}$}&{${mAP}_{AC}$}& &{$P_{AC}$}&{$R_{AC}$}&{${F1}_{AC}$}&{${mAP}_{AC}$}& &{$P_{AC}$}&{$R_{AC}$}&{${F1}_{AC}$}&{${mAP}_{AC}$}\\ \specialrule{.2em}{.1em}{.1em}
          I3D\cite{carreira2017quo}$*$ & 14.3 & 1.3 & 2.1 & 15.2 & &12.7 & 1.9 & 2.9 & 21.4 & &14.9 & 2.0&3.1&20.3 \\
          CF \cite{tirupattur2021modeling}$*$ & 10.3 &1.0 &1.6 &15.8& &9.0&1.5&2.2&22.2& & 10.7&1.6&2.4&21.0 \\
          MLAD \cite{tirupattur2021modeling}{{\checkmark}}&19.3 &7.2 &8.9 &28.9& & 18.9& 8.9& 10.5& 35.7& &19.6&9.0&10.8&34.8\\
          MS-TCT \cite{dai2022ms}{{\checkmark}}& 26.3 & 15.5 &19.5 &30.7 & & 27.6& 18.4& 22.1& 37.6 & & 27.9 &18.3 &22.1 &36.4 \\ \midrule
          {\method}{{\checkmark}} & \bf 28.3 &\bf 26.1 &\bf 27.2 &\bf 32.0 & &\bf 30.0 &\bf 29.2 &\bf 29.6 &\bf 37.8 & &\bf 30.0 &\bf 29.1 &\bf 29.4 &\bf 36.7 \\ \specialrule{.2em}{.1em}{.1em}
    \end{tabular}
  \caption{Action detection results on {\char} dataset based on the action-conditional metrics \cite{tirupattur2021modeling}, {$P_{AC}$}, {$R_{AC}$}, {${F1}_{AC}$}, and {${mAP}_{AC}$}. {$\tau$} refers the temporal window size. The same as \cite{dai2022ms,tirupattur2021modeling}, both RGB and optical flow are used for obtaining the results. The {{\checkmark}} symbol highlights the transformer-based approaches, and $*$ indicates the results are taken from \cite{tirupattur2021modeling}.\vspace{-2mm}} 
  \label{tab:sota v2}
\end{table*}

We also evaluate the performance of our proposed method by action-conditional metrics including Action-Conditional Precision {$P_{AC}$}, Action-Conditional Recall {$R_{AC}$}, Action-Conditional F1-Score {${F1}_{AC}$}, and Action-Conditional Mean Average Precision {${mAP}_{AC}$}, as introduced in \cite{tirupattur2021modeling}. The aim of these metrics is to measure the ability of the network to learn both co-occurrence and temporal dependencies of different action classes.  The metrics are measured throughout a temporal window with a size of $\tau$. As shown by the results {on {\char}} in Table \ref{tab:sota v2}, the proposed method {\method} achieves state-of-the-art results on all action-conditional metrics, specifically, it improves the state-of-the-art results significantly on $R_{AC}$ and ${F1}_{AC}$ by $10.6\%$ and $7.7\%$, $10.8\%$ and $7.5\%$, and $10.8\%$ and $7.3\%$ where $\tau$ is 0, 20, and 40 respectively.

\begin{figure}[h]
  \includegraphics[width=1.0\linewidth]{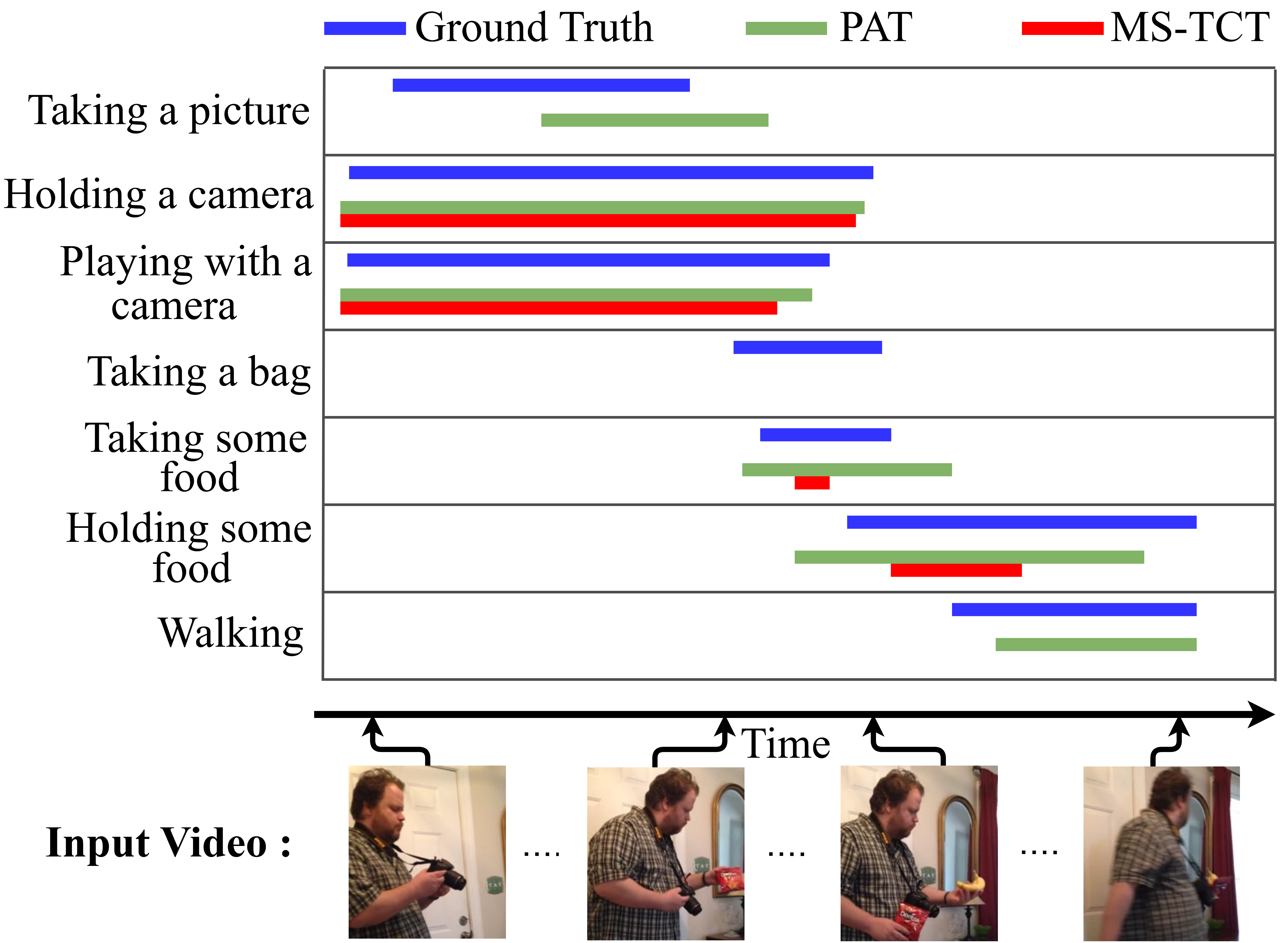}
  \caption{Visualization of action predictions by our proposed method {\method} and MS-TCT \cite{dai2022ms} on a test video sample of {\char} including 7 different action types. \vspace{-3mm}}
  \label{fig:qualitative}
\end{figure}

{Fig. \ref{fig:qualitative} displays qualitative results of {\method} on a test video sample of {\char} and compares them with the outputs of MS-TCT \cite{dai2022ms}. Amongst the state-of-the-art methods, we applied MS-TCT \cite{dai2022ms} and MLAD \cite{tirupattur2021modeling} on the video sample, since their code is available, useable and compatible with our hardware. However, as the MLAD could not predict any of the actions, we reported only the results of MS-TCT. The results in Fig. \ref{fig:qualitative} show that our proposed method’s action predictions have a better overlap with the ground-truth labels, and our method detected more action instances in the video than MS-TCT, {\ie} {\method} predicted all action types except “{\it{Taking a bag}}” while MS-TCT could not detect “{\it{Taking a picture}}”, “{\it Taking a bag}”, and “{\it{Walking}}”.}

\section{Conclusion}
\label{sec:conclusion}
In this work, we introduced a novel transformer-based network {\method} that exploits different ranges of temporal dependencies for action detection. The proposed method has been designed to benefit from preserving temporal positional information in learning multi-granularity features by (i) embedding the relative positional encoding in its transformer layers and (ii) a non-hierarchical design. We evaluated {\method} on two densely-labelled challenging benchmark action detection datasets, on which we achieved new state-of-the-art results, and our ablation studies demonstrated the effectiveness of different components of our proposed network. For future work, we will investigate adapting our network to learn spatial and temporal dependencies from raw pixels and also use audio information to improve the performance of action detection.

\section*{Acknowledgement}
This research is supported by UKRI EPSRC Platform Grant EP/P022529/1, and EPSRC BBC Prosperity Partnership AI4ME: Future Personalised Object-Based Media Experiences Delivered at Scale Anywhere EP/V038087/1.

{\small
\bibliographystyle{ieee_fullname}
\bibliography{egbib}

\begin{thebibliography}{40}
\providecommand{\natexlab}[1]{#1}
\providecommand{\url}[1]{\texttt{#1}}
\expandafter\ifx\csname urlstyle\endcsname\relax
  \providecommand{\doi}[1]{doi: #1}\else
  \providecommand{\doi}{doi: \begingroup \urlstyle{rm}\Url}\fi

\bibitem[Caba~Heilbron et~al.(2015)Caba~Heilbron, Escorcia, Ghanem, and
  Carlos~Niebles]{activitynet}
Fabian Caba~Heilbron, Victor Escorcia, Bernard Ghanem, and Juan Carlos~Niebles.
\newblock {ActivityNet: A large-Scale Video Benchmark for Human Activity
  Understanding}.
\newblock In \emph{Proceedings of the IEEE Conference on Computer Vision and
  Pattern Recognition}, pages 961--970, 2015.

\bibitem[Carreira and Zisserman(2017)]{carreira2017quo}
Joao Carreira and Andrew Zisserman.
\newblock {Quo vadis, Action Recognition? a New Model and the Kinetics
  Dataset}.
\newblock In \emph{proceedings of the IEEE Conference on Computer Vision and
  Pattern Recognition}, pages 6299--6308, 2017.

\bibitem[Chang et~al.(2021)Chang, Wang, Wang, Li, and Feng]{chang2021augmented}
Shuning Chang, Pichao Wang, Fan Wang, Hao Li, and Jiashi Feng.
\newblock {Augmented Transformer with Adaptive Graph for Temporal Action
  Proposal Generation}.
\newblock \emph{arXiv preprint arXiv:2103.16024}, 2021.

\bibitem[Chao et~al.(2018)Chao, Vijayanarasimhan, Seybold, Ross, Deng, and
  Sukthankar]{chao2018rethinking}
Yu-Wei Chao, Sudheendra Vijayanarasimhan, Bryan Seybold, David~A Ross, Jia
  Deng, and Rahul Sukthankar.
\newblock {Rethinking the Faster R-CNN Architecture for Temporal Action
  Localization}.
\newblock In \emph{proceedings of the IEEE Conference on Computer Vision and
  Pattern Recognition}, pages 1130--1139, 2018.

\bibitem[Dai et~al.(2021{\natexlab{a}})Dai, Das, and Bremond]{dai2021ctrn}
Rui Dai, Srijan Das, and Francois Bremond.
\newblock Ctrn: Class-temporal relational network for action detection.
\newblock \emph{British Machine Vision Conference}, 2021{\natexlab{a}}.

\bibitem[Dai et~al.(2021{\natexlab{b}})Dai, Das, Minciullo, Garattoni,
  Francesca, and Bremond]{dai2021pdan}
Rui Dai, Srijan Das, Luca Minciullo, Lorenzo Garattoni, Gianpiero Francesca,
  and Francois Bremond.
\newblock {PDAN: Pyramid Dilated Attention Network for Action Detection}.
\newblock In \emph{Proceedings of the IEEE Winter Conference on Applications of
  Computer Vision}, pages 2970--2979, 2021{\natexlab{b}}.

\bibitem[Dai et~al.(2022)Dai, Das, Kahatapitiya, Ryoo, and Bremond]{dai2022ms}
Rui Dai, Srijan Das, Kumara Kahatapitiya, Michael~S Ryoo, and Francois Bremond.
\newblock {MS-TCT: Multi-Scale Temporal ConvTransformer for Action Detection}.
\newblock In \emph{Proceedings of the IEEE Conference on Computer Vision and
  Pattern Recognition}, pages 20041--20051, 2022.

\bibitem[Dai et~al.(2019)Dai, Singh, Ng, and Davis]{dai2019tan}
Xiyang Dai, Bharat Singh, Joe Yue-Hei Ng, and Larry Davis.
\newblock {TAN: Temporal Aggregation Network for Dense Multi-Label Action
  Recognition}.
\newblock In \emph{2019 IEEE Winter Conference on Applications of Computer
  Vision}, pages 151--160. IEEE, 2019.

\bibitem[Devlin et~al.(2018)Devlin, Chang, Lee, and Toutanova]{devlin2018bert}
Jacob Devlin, Ming-Wei Chang, Kenton Lee, and Kristina Toutanova.
\newblock {BERT: Pre-Training of Deep Bidirectional Transformers for Language
  Understanding}.
\newblock \emph{arXiv preprint arXiv:1810.04805}, 2018.

\bibitem[Dosovitskiy et~al.(2020)Dosovitskiy, Beyer, Kolesnikov, Weissenborn,
  Zhai, Unterthiner, Dehghani, Minderer, Heigold, Gelly,
  et~al.]{dosovitskiy2020image}
Alexey Dosovitskiy, Lucas Beyer, Alexander Kolesnikov, Dirk Weissenborn,
  Xiaohua Zhai, Thomas Unterthiner, Mostafa Dehghani, Matthias Minderer, Georg
  Heigold, Sylvain Gelly, et~al.
\newblock {An Image Is Worth 16x16 Words: Transformers for Image Recognition at
  Scale}.
\newblock \emph{International Conference on Learning Representations}, 2020.

\bibitem[Dufter et~al.(2022)Dufter, Schmitt, and
  Sch{\"u}tze]{dufter2022position}
Philipp Dufter, Martin Schmitt, and Hinrich Sch{\"u}tze.
\newblock {Position Information in Transformers: An Overview}.
\newblock \emph{Computational Linguistics}, 48\penalty0 (3):\penalty0 733--763,
  2022.

\bibitem[Feichtenhofer(2020)]{feichtenhofer2020x3d}
Christoph Feichtenhofer.
\newblock {X3D: Expanding Architectures for Efficient Video Recognition}.
\newblock In \emph{Proceedings of the IEEE Conference on Computer Vision and
  Pattern Recognition}, pages 203--213, 2020.

\bibitem[Huang et~al.(2019)Huang, Vaswani, Uszkoreit, Shazeer, Simon,
  Hawthorne, Dai, Hoffman, Dinculescu, and Eck]{huang2018music}
Cheng-Zhi~Anna Huang, Ashish Vaswani, Jakob Uszkoreit, Noam Shazeer, Ian Simon,
  Curtis Hawthorne, Andrew~M Dai, Matthew~D Hoffman, Monica Dinculescu, and
  Douglas Eck.
\newblock {Music Transformer}.
\newblock \emph{International Conference on Learning Representations}, 2019.

\bibitem[Jiang et~al.(2014)Jiang, Liu, Zamir, Toderici, Laptev, Shah, and
  Sukthankar]{jiang2014thumos}
Yu-Gang Jiang, Jingen Liu, A~Roshan Zamir, George Toderici, Ivan Laptev,
  Mubarak Shah, and Rahul Sukthankar.
\newblock {THUMOS Challenge: Action Recognition with a Large Number of
  Classes}, 2014.

\bibitem[Kahatapitiya and Ryoo(2021)]{kahatapitiya2021coarse}
Kumara Kahatapitiya and Michael~S Ryoo.
\newblock {Coarse-Fine Networks for Temporal Activity Detection in Videos}.
\newblock In \emph{Proceedings of the IEEE Conference on Computer Vision and
  Pattern Recognition}, pages 8385--8394, 2021.

\bibitem[Kay et~al.(2017)Kay, Carreira, Simonyan, Zhang, Hillier,
  Vijayanarasimhan, Viola, Green, Back, Natsev, et~al.]{kay2017kinetics}
Will Kay, Joao Carreira, Karen Simonyan, Brian Zhang, Chloe Hillier, Sudheendra
  Vijayanarasimhan, Fabio Viola, Tim Green, Trevor Back, Paul Natsev, et~al.
\newblock {The Kinetics Human Action Video Dataset}.
\newblock \emph{arXiv preprint arXiv:1705.06950}, 2017.

\bibitem[Kingma and Ba(2015)]{kingma2014adam}
Diederik~P Kingma and Jimmy Ba.
\newblock {Adam: A Method for Stochastic Optimization}.
\newblock \emph{International Conference on Learning Representations}, 2015.

\bibitem[Li et~al.(2021)Li, Si, Li, Hsieh, and Bengio]{li2021learnable}
Yang Li, Si~Si, Gang Li, Cho-Jui Hsieh, and Samy Bengio.
\newblock {Learnable Fourier Features for Multi-Dimensional Spatial Positional
  Encoding}.
\newblock \emph{Advances in Neural Information Processing Systems},
  34:\penalty0 15816--15829, 2021.

\bibitem[Li and Yao(2021)]{li2021three}
Zhihui Li and Lina Yao.
\newblock {Three Birds with One Stone: Multi-Task Temporal Action Detection via
  Recycling Temporal Annotations}.
\newblock In \emph{Proceedings of the IEEE Conference on Computer Vision and
  Pattern Recognition}, pages 4751--4760, 2021.

\bibitem[Lin et~al.(2021)Lin, Xu, Luo, Wang, Tai, Wang, Li, Huang, and
  Fu]{lin2021learning}
Chuming Lin, Chengming Xu, Donghao Luo, Yabiao Wang, Ying Tai, Chengjie Wang,
  Jilin Li, Feiyue Huang, and Yanwei Fu.
\newblock {Learning Salient Boundary Feature for Anchor-Free Temporal Action
  Localization}.
\newblock In \emph{Proceedings of the IEEE Conference on Computer Vision and
  Pattern Recognition}, pages 3320--3329, 2021.

\bibitem[Lin et~al.(2018)Lin, Zhao, Su, Wang, and Yang]{lin2018bsn}
Tianwei Lin, Xu~Zhao, Haisheng Su, Chongjing Wang, and Ming Yang.
\newblock {BSN: Boundary Sensitive Network for Temporal Action Proposal
  Generation}.
\newblock In \emph{Proceedings of the European Conference on Computer Vision},
  pages 3--19, 2018.

\bibitem[Lin et~al.(2019)Lin, Liu, Li, Ding, and Wen]{lin2019bmn}
Tianwei Lin, Xiao Liu, Xin Li, Errui Ding, and Shilei Wen.
\newblock {BMN: Boundary-Matching Network for Temporal Action Proposal
  Generation}.
\newblock In \emph{Proceedings of the IEEE International Conference on Computer
  Vision}, pages 3889--3898, 2019.

\bibitem[Liu et~al.(2019)Liu, Ma, Zhang, Liu, and Chang]{liu2019multi}
Yuan Liu, Lin Ma, Yifeng Zhang, Wei Liu, and Shih-Fu Chang.
\newblock {Multi-Granularity Generator for Temporal Action Proposal}.
\newblock In \emph{Proceedings of the IEEE Conference on Computer Vision and
  Pattern Recognition}, pages 3604--3613, 2019.

\bibitem[Liu et~al.(2021)Liu, Lin, Cao, Hu, Wei, Zhang, Lin, and
  Guo]{liu2021swin}
Ze~Liu, Yutong Lin, Yue Cao, Han Hu, Yixuan Wei, Zheng Zhang, Stephen Lin, and
  Baining Guo.
\newblock {Swin Transformer: Hierarchical Vision Transformer Using Shifted
  Windows}.
\newblock In \emph{Proceedings of the IEEE International Conference on Computer
  Vision}, pages 10012--10022, 2021.

\bibitem[Ma et~al.(2020)Ma, Zhu, Yang, Zha, Kundu, Feiszli, and Shou]{ma2020sf}
Fan Ma, Linchao Zhu, Yi~Yang, Shengxin Zha, Gourab Kundu, Matt Feiszli, and
  Zheng Shou.
\newblock {SF-Net: Single-Frame Supervision for Temporal Action Localization}.
\newblock In \emph{Proceedings of the European Conference on Computer Vision},
  pages 420--437, 2020.

\bibitem[Piergiovanni and Ryoo(2019)]{piergiovanni2019temporal}
AJ~Piergiovanni and Michael Ryoo.
\newblock {Temporal Gaussian Mixture Layer for Videos}.
\newblock In \emph{International Conference on Machine learning}, pages
  5152--5161. PMLR, 2019.

\bibitem[Piergiovanni and Ryoo(2018)]{piergiovanni2018learning}
AJ~Piergiovanni and Michael~S Ryoo.
\newblock {Learning Latent Super-Events to Detect Multiple Activities in
  Videos}.
\newblock In \emph{Proceedings of the IEEE Conference on Computer Vision and
  Pattern Recognition}, pages 5304--5313, 2018.

\bibitem[Ridnik et~al.(2021)Ridnik, Ben-Baruch, Zamir, Noy, Friedman, Protter,
  and Zelnik-Manor]{ridnik2021asymmetric}
Tal Ridnik, Emanuel Ben-Baruch, Nadav Zamir, Asaf Noy, Itamar Friedman, Matan
  Protter, and Lihi Zelnik-Manor.
\newblock {Asymmetric Loss For Multi-Label Classification}.
\newblock In \emph{Proceedings of the IEEE International Conference on Computer
  Vision}, pages 82--91. IEEE Computer Society, 2021.

\bibitem[Shaw et~al.(2018)Shaw, Uszkoreit, and Vaswani]{shaw2018self}
Peter Shaw, Jakob Uszkoreit, and Ashish Vaswani.
\newblock {Self-Attention with Relative Position Representations}.
\newblock \emph{Proceedings of the Conference of the North American Chapter of
  the Association for Computational Linguistics: Human Language Technologies},
  2, 2018.

\bibitem[Shen et~al.(2018)Shen, Zhou, Long, Jiang, Pan, and
  Zhang]{shen2018disan}
Tao Shen, Tianyi Zhou, Guodong Long, Jing Jiang, Shirui Pan, and Chengqi Zhang.
\newblock {DiSAN: Directional Self-Attention Network for RNN/CNN-Free Language
  Understanding}.
\newblock In \emph{Proceedings of the AAAI Conference on Artificial
  Intelligence}, volume~32, 2018.

\bibitem[Sigurdsson et~al.(2016)Sigurdsson, Varol, Wang, Farhadi, Laptev, and
  Gupta]{charades}
Gunnar~A Sigurdsson, G{\"u}l Varol, Xiaolong Wang, Ali Farhadi, Ivan Laptev,
  and Abhinav Gupta.
\newblock {Hollywood in Homes: Crowdsourcing Data Collection for Activity
  Understanding}.
\newblock In \emph{Proceedings of the European Conference on Computer Vision},
  pages 510--526, 2016.

\bibitem[Tan et~al.(2022)Tan, Zhao, Shi, Kang, and Wang]{tanpointtad}
Jing Tan, Xiaotong Zhao, Xintian Shi, Bin Kang, and Limin Wang.
\newblock {PointTAD: Multi-Label Temporal Action Detection with Learnable Query
  Points}.
\newblock In \emph{Advances in Neural Information Processing Systems}, 2022.

\bibitem[Tirupattur et~al.(2021)Tirupattur, Duarte, Rawat, and
  Shah]{tirupattur2021modeling}
Praveen Tirupattur, Kevin Duarte, Yogesh~S Rawat, and Mubarak Shah.
\newblock {Modeling Multi-Label Action Dependencies for Temporal Action
  Localization}.
\newblock In \emph{Proceedings of the IEEE Conference on Computer Vision and
  Pattern Recognition}, pages 1460--1470, 2021.

\bibitem[Vahdani and Tian(2022)]{vahdani2022deep}
Elahe Vahdani and Yingli Tian.
\newblock {Deep Learning-based Action Detection in Untrimmed Videos: A Survey}.
\newblock \emph{IEEE Transactions on Pattern Analysis and Machine
  Intelligence}, 2022.

\bibitem[Vaswani et~al.(2017)Vaswani, Shazeer, Parmar, Uszkoreit, Jones, Gomez,
  Kaiser, and Polosukhin]{vaswani2017attention}
Ashish Vaswani, Noam Shazeer, Niki Parmar, Jakob Uszkoreit, Llion Jones,
  Aidan~N Gomez, {\L}ukasz Kaiser, and Illia Polosukhin.
\newblock {Attention Is All You Need}.
\newblock \emph{Advances in Neural Information Processing Systems}, 30, 2017.

\bibitem[Wang et~al.(2021)Wang, Xie, Li, Fan, Song, Liang, Lu, Luo, and
  Shao]{wang2021pyramid}
Wenhai Wang, Enze Xie, Xiang Li, Deng-Ping Fan, Kaitao Song, Ding Liang, Tong
  Lu, Ping Luo, and Ling Shao.
\newblock {Pyramid Vision Transformer: A Versatile Backbone for Dense
  Prediction without Convolutions}.
\newblock In \emph{Proceedings of the IEEE International Conference on Computer
  Vision}, pages 568--578, 2021.

\bibitem[Xu et~al.(2017)Xu, Das, and Saenko]{xu2017r}
Huijuan Xu, Abir Das, and Kate Saenko.
\newblock {R-C3D: Region Convolutional 3D Network for Temporal Activity
  Detection}.
\newblock In \emph{Proceedings of the IEEE International Conference on Computer
  Vision}, pages 5783--5792, 2017.

\bibitem[Xu et~al.(2020)Xu, Zhao, Rojas, Thabet, and Ghanem]{xu2020g}
Mengmeng Xu, Chen Zhao, David~S Rojas, Ali Thabet, and Bernard Ghanem.
\newblock {G-TAD: Sub-Graph Localization for Temporal Action Detection}.
\newblock In \emph{Proceedings of the IEEE Conference on Computer Vision and
  Pattern Recognition}, pages 10156--10165, 2020.

\bibitem[Yeung et~al.(2018)Yeung, Russakovsky, Jin, Andriluka, Mori, and
  Fei-Fei]{multithomus}
Serena Yeung, Olga Russakovsky, Ning Jin, Mykhaylo Andriluka, Greg Mori, and
  Li~Fei-Fei.
\newblock {Every Moment Counts: Dense Detailed Labeling of Actions in Complex
  Videos}.
\newblock \emph{International Journal of Computer Vision}, 126\penalty0
  (2):\penalty0 375--389, 2018.

\bibitem[Zhang et~al.(2022)Zhang, Wu, and Li]{zhang2022actionformer}
Chenlin Zhang, Jianxin Wu, and Yin Li.
\newblock {ActionFormer: Localizing Moments of Actions with Transformers}.
\newblock \emph{Proceedings of the European Conference on Computer Vision},
  2022.

\end{thebibliography}
}

\end{document}